\PassOptionsToPackage{numbers,sort&compress}{natbib}
\documentclass{article}

\usepackage[preprint]{neurips_2026}

\usepackage[utf8]{inputenc}
\usepackage[T1]{fontenc}
\usepackage{hyperref}
\usepackage{url}
\usepackage{booktabs}
\usepackage{amsmath, amssymb, amsfonts, amsthm}
\usepackage{nicefrac}
\usepackage{microtype}
\usepackage{xcolor}
\usepackage{graphicx}
\usepackage{array}
\usepackage{multirow}
\usepackage{calc}
\usepackage{caption}
\usepackage{float}
\usepackage{enumitem}
\usepackage{tabularx}
\usepackage{placeins}

\DeclareUnicodeCharacter{1D706}{\ensuremath{\lambda}}
\DeclareUnicodeCharacter{1D6FC}{\ensuremath{\alpha}}
\DeclareUnicodeCharacter{1D6FD}{\ensuremath{\beta}}
\DeclareUnicodeCharacter{1D6FE}{\ensuremath{\gamma}}
\DeclareUnicodeCharacter{1D6FF}{\ensuremath{\delta}}
\DeclareUnicodeCharacter{1D700}{\ensuremath{\epsilon}}
\DeclareUnicodeCharacter{1D702}{\ensuremath{\eta}}
\DeclareUnicodeCharacter{1D703}{\ensuremath{\theta}}
\DeclareUnicodeCharacter{1D705}{\ensuremath{\kappa}}
\DeclareUnicodeCharacter{1D707}{\ensuremath{\mu}}
\DeclareUnicodeCharacter{1D70C}{\ensuremath{\rho}}
\DeclareUnicodeCharacter{1D70E}{\ensuremath{\sigma}}
\DeclareUnicodeCharacter{1D70F}{\ensuremath{\tau}}
\DeclareUnicodeCharacter{1D714}{\ensuremath{\omega}}
\DeclareUnicodeCharacter{03BB}{\ensuremath{\lambda}}
\DeclareUnicodeCharacter{03B1}{\ensuremath{\alpha}}
\DeclareUnicodeCharacter{03B2}{\ensuremath{\beta}}
\DeclareUnicodeCharacter{03C9}{\ensuremath{\omega}}

\title{L-FNO: Lorentzian Fourier Neural Operator for Stochastic Event Dynamics}

\author{%
  Jihoon Kang\thanks{Associate Professor}, Songhee~Kang\thanks{Assistant Professor, Corresponding Author} \\
  Department of Business Administration\\
  Tech University of Korea\\
  Sangidaehak-ro 237, Siheung-si, Gyeonggi-do, Republic of Korea \\
  \texttt{dellabee@tukorea.ac.kr} \\
}

\begin{document}

\maketitle

\begin{abstract}
Modern operational systems face uncertainty even in routine conditions, where rare, bursty, and self-exciting events emerge from both exogenous covariates and endogenous event dynamics. Standard neural operators are typically trained as regression-style function-to-function models rather than conditional-intensity estimators, limiting their suitability for sparse event regimes. We introduce the Lorentzian Fourier Neural Operator (L-FNO), a stochastic neural operator that combines an FNO-style covariate path, Lorentzian spectral kernels for history-dependent excitation, and a likelihood-based training objective. We evaluate L-FNO on eight synthetic point-process benchmarks and three real-world datasets covering disease outbreak prediction and semiconductor fault or defect detection. L-FNO improves event likelihood, calibration diagnostics, and rare-event detection over regression- and likelihood-based neural operator baselines. These results show that structured spectral memory and likelihood-based learning provide effective inductive biases for neural operator models of stochastic event dynamics.
\end{abstract}

\section{Introduction}

High-impact events, which often pose critical operational risks, such as semiconductor defects, financial fraud, and epidemics, seem rare, irregular, and unpredictable. However, these events exhibit a nontrivial temporal structure, including clustering and context-dependent bursts. The key research questions are (i) whether such a structure can be learned from inherently sparse data and (ii) whether neural operator frameworks can, in principle, capture it.

Regarding the first question, this study confirmed that structural patterns can be learned by combining potential event dependencies with exogenous variables, even when the data are sparse. However, to the second question, our answer is cautious: not without modification. Existing neural operator frameworks lack an explicit mechanism of predictability under uncertainty, which is essential when observations are sparse and outcomes are stochastic. Neural operators such as the Fourier Neural Operator (FNO) \citep{li2021fno} and DeepONet \citep{lu2021deeponet} have achieved strong performance in surrogate modeling of Partial Differential Equations (PDEs) \citep{yao2025difno}, where targets are smooth and largely deterministic. However, applying this regression framework to rare events holds a critical misalignment with the core problem definition, since the predictor corresponds to the conditional mean of stochastic event processes. This approach fails to capture the likelihood structure of discrete events, resulting in poor calibration, particularly for rare or extreme outcomes. Even weighted mean-squared error baselines designed to emphasize rare events substantially degrade probabilistic calibration, often by orders of magnitude across benchmarks.

Temporal Point Processes (TPPs) provide a more principled approach to modeling these events. The Hawkes process \citep{hawkes1971spectra} models self-excitation through an exponential kernel. That means each past event raises the probability of near-future events by an amount that decays over time. Neural extensions \citep{mei2017neuralhawkes,zhang2021universal,zuo2020thp} offer greater flexibility, but their focus on event sequences limits their ability to incorporate rich exogenous covariates. It is a critical gap in applications such as epidemic forecasting, where climate and population data drive outbreak risk.

\begin{figure}[t]
  \centering
  \includegraphics[width=5.5in,height=2.2in]{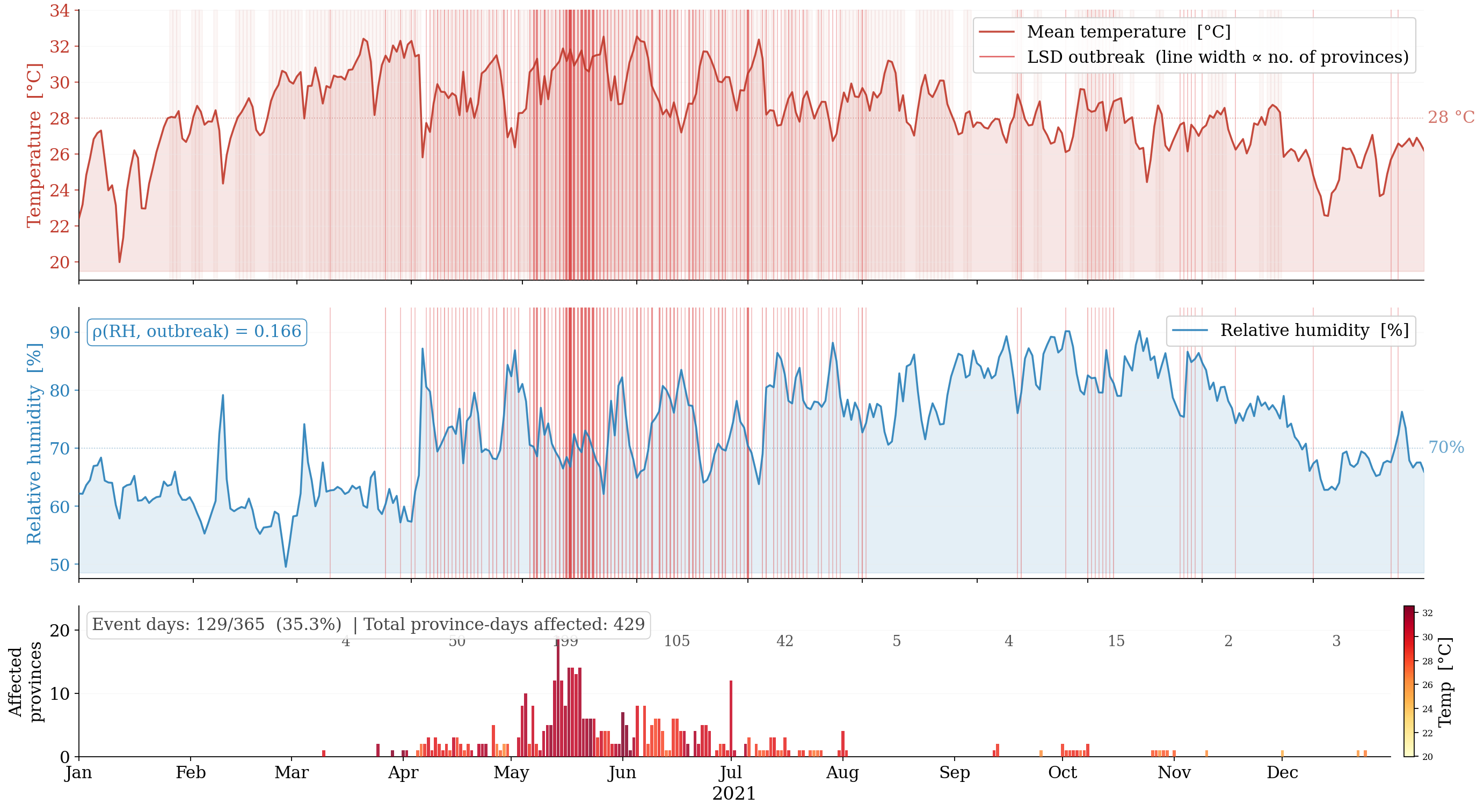}
  \caption{Lumpy Skin Disease (LSD) outbreaks in Thailand (2021), a motivating example.}
  \label{fig:1}
  \vspace{0.25em}
  \begin{minipage}{5in}
    \scriptsize
    \textit{Note.} (top) Daily national mean temperature and (middle) relative humidity across 67 provinces. Red vertical lines mark days with confirmed LSD outbreak events; line width is proportional to the number of affected provinces on that day. (bottom) Daily count of affected provinces, colored by temperature. Outbreaks are heavily concentrated during the hot, humid season (April--June), yet the mapping from climate covariates to outbreak events is neither linear nor instantaneous: long quiescent periods are punctuated by dense cascade clusters driven by wind-borne transmission of infectious agents. Due to the low event rate (1.8\%) of outbreak data and the lack of reliable climate thresholds, a joint covariate history model is required to capture both the climate-driven baselines and self-exciting dynamics in disease transmission.
  \end{minipage}
\end{figure}

This paper proposes the Lorentzian Fourier Neural Operator (L-FNO) to address the research gap. The L-FNO extends FNO architecture with a Lorentzian spectral kernel that corresponds exactly to an exponential excitation function, a hallmark of Hawkes dynamics. This parameterization is not only convenient: Proposition 1 describes that the Lorentzian transfer function \(\alpha_{k}/(\beta_{k}+i\omega)\) is the Fourier transform of the exponential kernel \(\alpha_{k}e^{-\beta_k t}\), establishing a direct equivalence between L-FNO and the self-excitation term of Hawkes process. Trained under a Poisson negative log-likelihood loss (NLL), L-FNO learns an event intensity function, modeling both covariate-driven effects and history-dependent dynamics within a single operator.

The main contributions of this research are summarized in three folds.

First, L-FNO extends the current FNO architecture by incorporating a structured Lorentzian spectral kernel, \emph{\ensuremath{\sum_k} \ensuremath{\alpha_k}/(\ensuremath{\beta_k} + i\ensuremath{\omega}).} By the convolution theorem, this kernel is equivalent in the time domain to a multi-scale Hawkes excitation function, \emph{\ensuremath{\sum_k}} \(\alpha_ke^{- \beta_kt}\), providing a spectral formulation of self-exciting point process dynamics. The paper further shows that unconstrained signed or complex Lorentzian mixtures are dense in the Hardy space \(H^2(\mathbb{C}_+)\) \citep{feichtinger2013hyperbolic,garcia2016modelspaces}, establishing an expressivity result for general causal excitation kernels (\emph{Proposition 1}). The positive-amplitude L-FNO implementation used in this paper targets the excitatory Hawkes setting and therefore approximates a restricted class of nonnegative causal excitation kernels.
Second, the paper characterizes a practical failure mode of regression-based MSE operator learning under sparse event observations. In the intercept-only case, or when the event signal is weak relative to the majority non-event gradient, the MSE solution is biased toward the marginal event rate \(p\), producing near-constant predictions with limited temporal sensitivity \citep{hastie2009elements}. Event upweighting can partially increase sensitivity to rare positives, but it trades off probabilistic calibration for marginal detection gain \citep{caplin2022calibrating,gneiting2007proper,he2009imbalanced,murphy1973probabilityscore}. Training the model under a Poisson NLL objective resolves this objective mismatch by providing asymmetric gradient pressure at both event and non-event timesteps.
Third, the learned Lorentzian parameters (\emph{\ensuremath{\alpha_k}} and \emph{\ensuremath{\beta_k}}) serve as an interpretable model-driven measure of a system's history-dependent predictability structure: \emph{\ensuremath{\alpha_k}} quantifies the strength of self-excitation and \(\tau_k = 1/\beta_k\) its temporal persistence. Therefore, L-FNO moves the current academic discourse beyond event prediction to the quantification of predictive temporal structure embedded in a stochastic event system, drawing a clear distinction from conventional neural operators and point process models.
\section{Related Work}

\subsection{Neural Operators for Functional-to-Functional Learning}

Neural operators learn mappings between function spaces in a discretization-invariant manner, typically from exogenous input fields to solution fields \citep{kovachki2023neuraloperator}. Representative approaches include DeepONet and Fourier Neural Operator (FNO), which enable scalable surrogate modeling for parametric PDEs \citep{li2021fno,lu2021deeponet}. These methods are specifically relevant in settings with high-dimensional, structured inputs, which aligns with our use of exogenous fields rather than tabular covariates. Subsequent variants have improved representation efficiency, multiresolution structure, and trainability \citep{kovachki2023neuraloperator,li2021fno,tripura2022wavelet}. However, neural operators are still primarily designed to predict continuous fields or trajectories. They are not naturally formulated as conditional intensity models for discontinuous stochastic events.

\subsection{Temporal Point Processes for Continuous-Time Stochastic Event Modeling}

Temporal point processes (TPPs) provide a probabilistic framework for modeling stochastic event arrivals in continuous time \citep{daley2003pointprocesses,hawkes1971spectra}. Classical Hawkes processes model self-excitation through history-dependent intensity functions, while neural TPPs replace hand-specified kernels with learned sequence representations. Early neural models such as RMTPP connect recurrent neural networks with marked temporal point processes \citep{du2016rmtpp}. Neural Hawkes Process, SAHP, and THP further parameterize conditional intensities or next-event distributions using continuous-time recurrent dynamics or attention-based history encoders \citep{mei2017neuralhawkes,zhang2020sahp,zuo2020thp}. Other neural TPP formulations improve likelihood evaluation, generative modeling, or hybrid continuous-discrete dynamics \citep{omi2019fully,xiao2017wasserstein,jia2019neuraljump}. Extensions to spatiotemporal settings further enhance flexibility \citep{chen2021neuralstpp}.

Despite these strengths, most neural TPPs are primarily history-driven sequence models \citep{shchur2021neural}. They may incorporate marks or auxiliary features, but they do not natively learn resolution-invariant mappings from high-dimensional exogenous fields to stochastic event distributions. This distinction is central to our setting, where external functional covariates, rather than event history alone, can trigger rare events.

\subsection{Rare and Extreme Event Prediction Beyond Imbalanced Classification}

Tailored learning strategies are required when predicting rare events in the face of extreme imbalance and tail risk \citep{shyalika2024rareeventsurvey}. In dynamical systems, extreme events are often linked to instabilities and are studied through combined mechanistic and data-driven approaches \citep{chowdhury2022extremeevents}. However, most formulations reduce the problem to classification, anomaly detection, or threshold exceedance, compressing events into sparse labels or risk scores. This abstraction obscures generative mechanisms, conditional intensity structure, and continuous-time event dynamics.

\subsection{Gap and Positioning: Operator Learning for Stochastic Rare Event Dynamics}

Existing work addresses functional conditioning, stochastic event modeling, and rare-event prediction largely in isolation. Neural operators provide a natural language for mappings between high-dimensional fields, but their outputs are usually continuous solution fields rather than event distributions. Neural TPPs provide likelihood-based models for stochastic events, but they are typically built around endogenous event histories rather than exogenous functional inputs. Rare-event methods address class imbalance and tail risk, but often collapse event dynamics into binary or thresholded labels.

Our work is positioned at this intersection. We propose an operator-learning perspective in which the goal is to learn a mapping from functional covariates and event history to stochastic event intensities. The proposed L-FNO combines an FNO-style exogenous covariate path with a Lorentzian event-history memory path, retaining sensitivity to high-dimensional external conditions while modeling continuous-time stochastic event generation under rare and extreme regimes.

\section{Method}

\subsection{Problem Formulation}

Let \(dN(t)\in\{0,1\}\) denote the observed event indicator in a unit-width discrete time bin, and let \(X(t)\in\mathbb{R}^{k}\) be a vector of exogenous covariates observed at the same time. Define the event history up to time \(t\) as
\[
H_t=\{dN(s):s<t\}.
\]
The goal is to learn the conditional event intensity
\[
\lambda(t)>0,
\]
which parameterizes the expected number of events in a unit-width time bin conditional on \(X(t)\) and \(H_t\).

Although the observed event bins are binary in our experiments, we model them as unit-width binned counts under a Poisson observation model. Thus, \(\lambda(t)\) denotes a conditional intensity, not a Bernoulli probability. For a unit-width bin, the probability of at least one event is \(1-\exp[-\lambda(t)]\), while the Poisson negative log-likelihood for the observed count \(dN(t)\in\{0,1\}\), up to constants independent of the model parameters, is
\[
\mathcal{L}_{\mathrm{NLL}}
=
\mathbb{E}_{t}
\left[
\lambda(t)-dN(t)\log\lambda(t)
\right].
\]
This formulation aligns the learning objective with the conditional-intensity view of discretized point processes while retaining binary event labels for detection metrics.

\subsection{Lorentzian Fourier Neural Operator}

L-FNO learns \emph{λ(t)} through three L-FNO Blocks stacked in sequence, each of which processes both \emph{X(t)} and the event history \emph{h(t)} in the spectral domain. The full intensity estimate is

\begin{equation}
\widehat{\lambda}(\omega)
=
R(\omega)\cdot\widehat{x}(\omega)
+
\sum_{k}
\left[
\frac{\alpha_k}{\beta_k+i\omega}
\right]
\cdot\widehat{h}(\omega)
\label{eq:lfno_spectral}
\end{equation}

\begin{equation}
\lambda(t)
=
\operatorname{Softplus}
\left(
\mathcal{F}^{-1}
\left[
\widehat{\lambda}(\omega)
\right]
\right)
\label{eq:lfno_intensity}
\end{equation}
where $\widehat{x}(\omega)$ and $\widehat{h}(\omega)$ are the discrete Fourier transforms of the covariate sequence $X(t)$ and the strictly lagged event-history sequence $h(t)$, respectively. The Lorentzian factor $\alpha_k/(\beta_k+i\omega)$ provides a learnable spectral weighting of $\widehat{h}(\omega)$ and induces the exponential-memory response in the time domain. Here, $R(\omega)$ is a learnable complex-valued operator applied to $K$ low-frequency modes, and $\alpha_k$ and $\beta_k$ are per-channel amplitude and decay parameters. Each L-FNO Block applies this transformation, followed by a skip connection and GELU nonlinearity.
\begin{figure}[t]
  \centering
  \includegraphics[width=3.5in,height=2.5in]{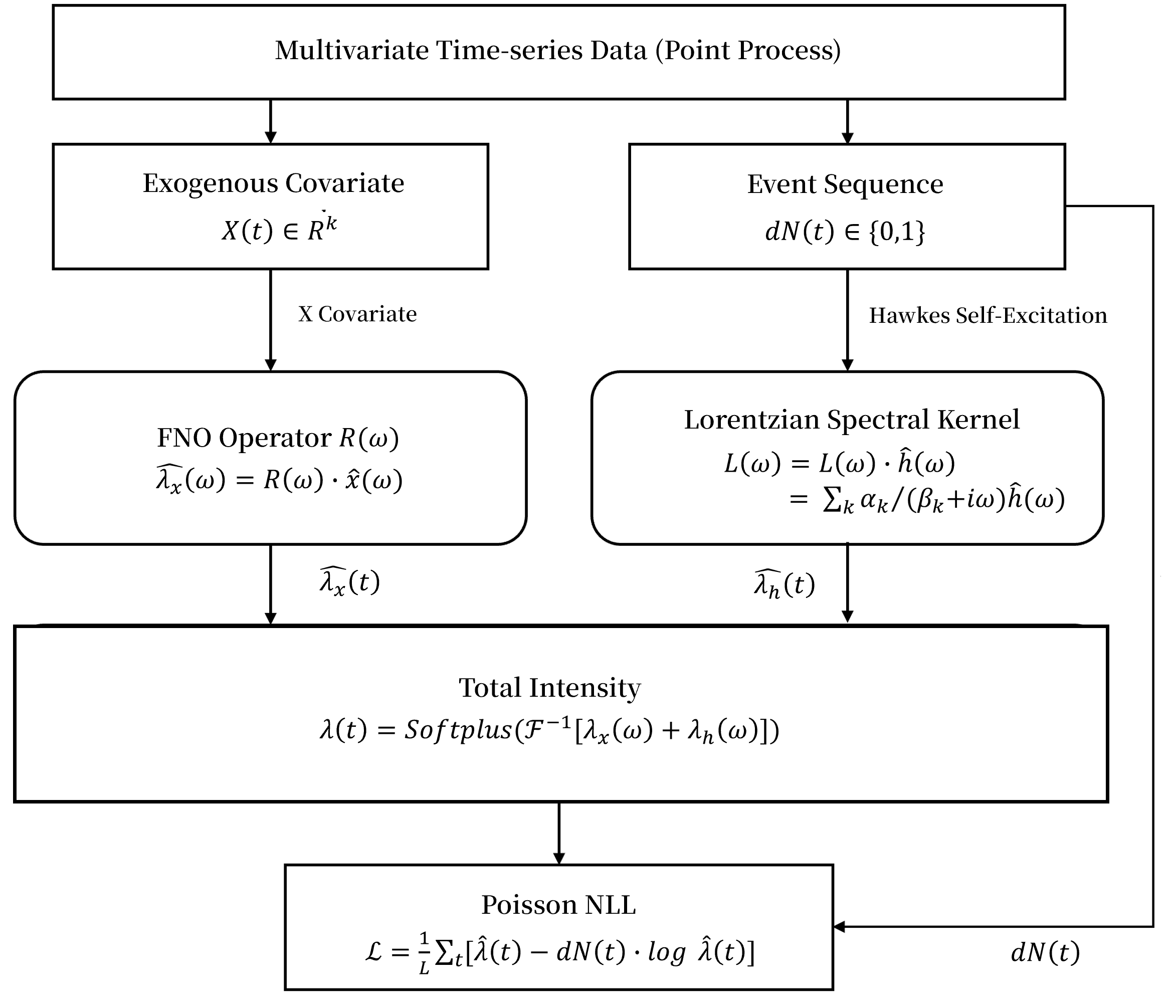}
  \caption{Overview of the L-FNO architecture for modeling stochastic event dynamics.}
  \label{fig:2}
  \vspace{0.25em}
  \begin{minipage}{5in}
    \scriptsize
    \textit{Note.} L-FNO combines an FNO-based exogenous covariate path with a Lorentzian event-history memory block. The Lorentzian block is motivated by its Fourier--Laplace transfer function but evaluated as a one-sided causal exponential recurrence, enabling joint learning of baseline intensities and history-dependent excitation without future-event access.  \end{minipage}
\end{figure}

\subsection{Lorentzian Spectral Kernel}

The term \(\alpha_k/(\beta_k+i\omega)\) is the Fourier transform of the causal exponential function
\[
g_k(t)=\alpha_k e^{-\beta_k t}\mathbf{1}_{\{t\geq 0\}}.
\]
In the time domain, the stochastic component of the L-FNO output is therefore
\[
\varphi(t)
=
h(t) *
\left(
\sum_k \alpha_k e^{-\beta_k t}\mathbf{1}_{\{t\geq 0\}}
\right).
\]

This is a sum of exponential excitations driven by the strictly lagged event-history impulse signal. Since \(h(t)\) contains only past events, the resulting convolution implements a causal Hawkes-type excitation kernel \citep{hawkes1971spectra}.

\begin{quote}
\emph{\textbf{Proposition 1 (Approximation of causal kernels by stable exponential modes).} Let} 
\(g \in L^{2}\left([0,\infty)\right)\) 
\emph{be a causal finite-energy kernel. Then for any} 
\(\varepsilon > 0\), 
\emph{there exist an integer} \(K\), 
\emph{coefficients} 
\(\alpha_{1},\ldots,\alpha_{K}\in\mathbb{C}\), 
\emph{and decay rates} 
\(\beta_{1},\ldots,\beta_{K} > 0\) 
\emph{such that}
\[
\left\| 
g(t) - \sum_{k = 1}^{K}\alpha_{k}e^{-\beta_k t}
\right\|_{L^{2}\left([0,\infty)\right)} 
< \varepsilon.
\]
\emph{Equivalently, under the Paley--Wiener isometry, the corresponding stable rational transfer functions}
\[
\sum_{k = 1}^{K}\frac{\alpha_{k}}{\beta_{k} + s}
\]
\emph{are dense in} \(H^{2}(\mathbb{C}_{+})\), 
\emph{the Hardy space representation of causal} \(L^{2}\) 
\emph{kernels. Their boundary values on} \(s = i\omega\),
\[
\sum_{k = 1}^{K}\frac{\alpha_{k}}{\beta_{k} + i\omega},
\]
\emph{approximate the Fourier--Laplace boundary response of} \(g\) 
\emph{in} \(L^{2}(\mathbb{R}_{\omega})\). 
\emph{Each first-order pole has a Lorentzian magnitude-squared profile, while the complex response also encodes phase.}
\emph{When} \(g\) \emph{is real-valued, the coefficients can be chosen real-valued.}
\end{quote}

\textbf{Proposition 1} shows that finite signed or complex Lorentzian mixtures are dense in causal \(L^{2}\) kernels, equivalently in their \(H^{2}\) Fourier--Laplace responses. The implemented L-FNO uses positive amplitudes, so it represents the excitatory Hawkes subclass rather than arbitrary signed kernels. Increasing the number of poles therefore expands the available causal memory profiles within this excitatory class. The proof is given in Appendix~\ref{app:proof}.

\subsection{Regression Failure Under Sparse Events}

The core challenge in stochastic event prediction under sparse observations is not only model capacity, but also objective mismatch. The paper makes this precise through a simple decomposition. Let \(p=\mathbb{E}[dN(t)]\ll 1\). Under Mean Squared Error (MSE), the per-sample gradient with respect to the predicted intensity \(\widehat{\lambda}\) is
\[
\frac{\partial \mathcal{L}_{\mathrm{MSE}}}{\partial \widehat{\lambda}}
=
2\left(\widehat{\lambda} - dN(t)\right).
\label{eq:mse_gradient}
\]

Non-event timesteps (\(dN(t)=0\), fraction \(1-p\approx 1\)) push \(\widehat{\lambda}\) toward zero, while event timesteps (\(dN(t)=1\), fraction \(p\)) push \(\widehat{\lambda}\) upward. In an intercept-only model, these forces balance at the marginal event rate,
\[
\widehat{\lambda}_{\mathrm{MSE}}
\rightarrow
(1-p)\cdot 0 + p\cdot 1
=
p.
\]

More generally, the population MSE optimum is the conditional mean \(\mathbb{E}[dN(t)\mid X(t),H_t]\). Therefore, MSE is not theoretically restricted to constant prediction when informative covariates and histories are available. However, in sparse-event regimes, the overwhelming number of non-event timesteps produces a practical mean-collapse failure mode: gradient updates are dominated by zeros, and the learned predictor is biased toward low, weakly dynamic intensity estimates unless the event signal is sufficiently strong. This is the objective-level failure mode that motivates a likelihood-based formulation.

The Poisson NLL gradient,
\[
\frac{\partial \mathcal{L}_{\mathrm{NLL}}}{\partial \widehat{\lambda}}
=
1 - \frac{dN(t)}{\widehat{\lambda}},
\label{eq:nll_gradient}
\]

behaves asymmetrically. At event timesteps (\(dN(t)=1\)), the gradient diverges as \(\widehat{\lambda}\rightarrow 0\), exerting an unbounded upward pull that prevents severe under-prediction of rare events. At non-event timesteps (\(dN(t)=0\)), the gradient equals 1, maintaining steady downward pressure without the quadratic suppression induced by MSE. This two-sided structure provides the calibration signal that sparse event modeling requires and is absent from MSE by construction.

\subsection{Objective Function of L-FNO}

L-FNO minimizes Poisson NLL over a sliding window of length L:

\begin{equation}
\mathcal{L}_{\mathrm{NLL}}
=
\frac{1}{L}
\sum_{t=1}^{L}
\left[
\widehat{\lambda}(t)
-
dN(t)\log \widehat{\lambda}(t)
\right].
\end{equation}

where the predicted intensity is given by:
\begin{equation}
\widehat{\lambda}(t)
=
\operatorname{Softplus}
\left(
\mathcal{F}^{-1}
\left[
R(\omega)\widehat{x}(\omega)
+
\sum_{k=1}^{K}
\frac{\alpha_k}{\beta_k + i\omega}
\widehat{h}(\omega)
\right](t)
\right).
\end{equation}

This objective directly targets the event intensity function rather than a regression residual, and its gradient structure, as established in Section 3.4 provides calibration pressure at both event and non-event timesteps. Softplus output activation ensures \(\widehat{\lambda}(t) > 0\) throughout, avoiding numerical issues at event times. At every timestep (\(dN(t)\) = 1), the loss diverges as \(\widehat{\lambda}(t)\) = 0, penalizing under-prediction without bound. On the other hand, at non-event timesteps (\(dN(t)\) = 0), \(\mathcal{L}_{NLL} = \ \widehat{\lambda}\) provides a constant downward gradient which penalizes over-prediction proportionally. This asymmetric structure provides calibration pressure at both extremes, which is a property absent from any MSE-based objective by construction.

Taken together, the three design choices of L-FNO form a coherent whole. The Lorentzian spectral kernel encodes the self-exciting structure of point processes directly in the frequency domain (Section 3.3). The FNO operator stack handles exogenous covariate effects at arbitrary temporal scales (Section 3.2). The Poisson NLL objective aligns the learning signal with the generative process of the observed events (Section 3.4). Each component addresses a distinct limitation of applying standard neural operators to stochastic event tasks.

\section{Experiments}

\subsection{Synthetic Benchmarks}

To systematically evaluate the performance across diverse point-process scenarios, the study constructs eight synthetic benchmarks by simulating events from the intensity functions following the Hawkes process with varying structural properties. Table~\ref{tab:synthetic_scenarios} summarizes four characterization statistics for each scenario: the event rate, the Fano factor \citep{lowen1995fractalpointprocesses}, the Allan factor slope \citep{lowen1995fractalpointprocesses,lowen1996allanvariance}, and the memory coefficient \(M\) \citep{goh2008burstiness}. Together, these statistics identify whether a scenario is dominated by rarity, clustering, multi-scale self-excitation, or long-range time series pattern dependence.

\begin{table}[t]
\small
\setlength{\tabcolsep}{3.5pt}
\renewcommand{\arraystretch}{1.12}
\caption{Synthetic benchmark scenarios and point-process characteristics.}
\label{tab:synthetic_scenarios}
\begin{tabular*}{5.5in}{@{\extracolsep{\fill}}lrrrr}
\toprule
\textbf{Scenario} &
\textbf{Rate} &
\textbf{Fano} &
\textbf{AF Slope} &
\textbf{Memory} \\
\midrule
B1-Rare         & 0.039 & 2.652  & 0.208  & 0.024  \\
B2-Cascade      & 0.141 & 11.411 & 0.644  & 0.120  \\
B3-Burst        & 0.041 & 9.146  & 0.490  & -0.011 \\
B4-MultiScale   & 0.072 & 8.303  & 0.550  & 0.163  \\
B5-Nonlinear    & 0.162 & 4.953  & 0.437  & 0.099  \\
B6-Inhibitory   & 0.095 & 0.819  & -0.062 & 0.032  \\
B7-ZeroInflated & 0.046 & 13.968 & 0.601  & 0.021  \\
B8-LongMemory   & 0.027 & 2.171  & 0.172  & 0.093  \\
\bottomrule
\end{tabular*}
\end{table}

B1-Rare captures low-rate sparse events with mild clustering, representative of equipment faults or abnormal patterns in financial transactions. B2-Cascade models self-exciting events that trigger neighborhood spread, analogous to epidemic propagation. B3-Burst generates intense but short-lived activity clusters. B4-MultiScale combines fast and slow excitation timescales. B5-Nonlinear introduces threshold-dependent intensity. B6-Inhibitory models a sub-Poisson regime in which past events reduce future probabilities. B7-ZeroInflated generates long silent periods punctuated by multi-event bursts. Last, B8-LongMemory encodes dependencies spanning 50 or more timesteps.

\subsection{Real-World Datasets}

Three real-world datasets are selected to cover the overall range of mapped synthetic regimes. Table~\ref{tab:scenario_statistics} summarizes their characteristics with point-process statistics.

\begin{table}[htbp]
\centering
\small
\setlength{\tabcolsep}{3.5pt}
\renewcommand{\arraystretch}{1.12}
\caption{Real-world datasets and point-process characteristics.}
\label{tab:scenario_statistics}
\begin{tabularx}{5.5in}{@{}c l 
>{\centering\arraybackslash}p{0.50in}
>{\centering\arraybackslash}p{0.55in}
>{\centering\arraybackslash}p{0.50in}
>{\centering\arraybackslash}p{0.50in}
>{\raggedright\arraybackslash}X@{}}
\toprule
\textbf{Case} &
\textbf{Dataset} &
\textbf{\shortstack{Event Rate}} &
\textbf{\shortstack{Fano Factor}} &
\textbf{\shortstack{AF Slope}} &
\textbf{Memory} &
\textbf{Mapped Scenarios} \\
\midrule
1 & FDC   & 0.100 & 3.150 & 0.350 & 0.202 & Rare, zero-inflated, and burst \\
2 & SECOM & 0.066 & 2.520 & 0.270 & 0.171 & Rare \\
3 & LSD   & 0.018 & 5.000 & 0.380 & 0.067 & Rare, zero-inflated, burst, and cascade \\
\bottomrule
\end{tabularx}
\end{table}

The FDC case is a proprietary semiconductor fault-detection series, where events correspond to process anomalies exceeding the 90th percentile control limit in the etching process. The UCI SECOM dataset contains wafer-defect labels and sensor measurements from semiconductor manufacturing \citep{mccann2008secom}. The LSD dataset comprises daily province-level Lumpy Skin Disease outbreak records from Thailand in 2021, enriched with ERA5 climate covariates and a wind-directed spatial Hawkes history term capturing inter-province cascade dynamics \citep{woah_wahis_dashboard}. A comprehensive visualization of the results is provided in Appendix~\ref{app:visualizations}.

\subsection{Baselines and Evaluation Metrics}

The proposed L-FNO is compared against two FNO-based baselines that isolate the contribution of each design choice. FNO-MSE is a standard FNO trained with mean squared error, representing the conventional regression approach to event prediction. FNO-NLL shares the same backbone architecture and Poisson NLL objective as L-FNO but removes the Lorentzian recurrence, allowing the history representation to be learned by a generic operator path. For all FNO-based baselines, the event-history input is strictly lagged as \(h(t)=dN(t-1)\). FNO-NLL follows the same chronological input protocol as L-FNO; thus, the performance gap between FNO-NLL and L-FNO isolates the effect of the structured Lorentzian causal memory prior under the same likelihood objective. To connect with the temporal point process literature, Appendix~\ref{app:additional_baselines} reports an extended comparison with Neural Hawkes \citep{mei2017neuralhawkes} and Neural Hawkes with covariates (NH-X) \citep{meng2024transfeattpp,xiao2019learning}, which serve as recurrent neural TPP baselines under the same evaluation protocol.
All models are trained using AdamW with cosine annealing over 300 epochs, learning rate \(3\times10^{-4}\), and a sliding window of length \(L=64\)--96 timesteps. Hyperparameters are fixed across datasets to ensure a consistent comparison protocol.

Performance is reported on four complementary metrics: NLL, Brier score, PR-AUC, and AUC. NLL measures likelihood quality and serves as the primary probabilistic metric. Brier score is used as an auxiliary binary-label calibration diagnostic, computed as \(\mathrm{Brier}=N^{-1}\sum_t(\widehat{\lambda}(t)-dN(t))^2\) with \(dN(t)\in\{0,1\}\). PR-AUC is the primary rare-event detection metric, and AUC is reported as a complementary discrimination metric.
\section{Experimental Results}

The study distinguishes two baseline groups: operator ablations or loss-function variants, including FNO-MSE, FNO-NLL, and FNO-WMSE; and external temporal point process baselines, including Neural Hawkes (NH) and covariate-augmented Neural Hawkes (NH-X). The main results in Section 5 describe FNO-MSE and FNO-NLL, while full results for FNO-WMSE, NH, and NH-X are provided in Appendix~\ref{app:additional_baselines}. To isolate the performance gain, we distinguish between the stochastic objective (Poisson NLL) and the architectural prior (Lorentzian kernel). While FNO-NLL shares the same likelihood-based objective, L-FNO consistently outperforms it across the real-world benchmarks (Table~\ref{tab:real_world_results}). This performance gap demonstrates that the Lorentzian kernel acts as a critical inductive bias, enabling the explicit parameterization of self-exciting temporal structures.

\subsection{Synthetic Benchmarks}

Table~\ref{tab:synthetic_results} summarizes the multi-seed synthetic benchmark results by comparing L-FNO against the strongest baseline for each scenario and metric. Full results for all baselines and all metrics, including Brier score and AUC, are provided in Appendix~\ref{app:full_results}. L-FNO achieves the best PR-AUC in all eight synthetic scenarios and improves the macro-average PR-AUC from $0.347{\pm}0.085$ to $0.748{\pm}0.048$. It also improves the average NLL from $0.231{\pm}0.019$ for the best baseline to $0.190{\pm}0.019$.
The main exceptions are calibration-oriented metrics in selected regimes. FNO-NLL achieves the lowest NLL in B2-Cascade and B8-ZeroInflated, and the lowest Brier score in B2-Cascade, B7-ZeroInflated, and B8-LongMemory. This suggests that while the Lorentzian memory improves event ranking and rare-event discrimination, likelihood calibration can be more sensitive in regimes with frequent cascades, structurally silent periods, or long-memory dynamics.

\begin{table}[htbp]
\centering
\footnotesize
\setlength{\tabcolsep}{5pt}
\renewcommand{\arraystretch}{1.08}
\caption{Synthetic benchmark summary over five random seeds. We compare L-FNO against the best baseline for each scenario and metric. Full mean$\pm$std results for all metrics and baselines are reported in Appendix~\ref{app:full_results}.}
\label{tab:synthetic_results}
\begin{tabular*}{5.5in}{@{\extracolsep{\fill}}lcccc@{}}
\toprule
\multirow{2}{*}{\textbf{Scenario}} &
\multicolumn{2}{c}{\textbf{NLL} $\downarrow$} &
\multicolumn{2}{c}{\textbf{PR-AUC} $\uparrow$} \\
\cmidrule(lr){2-3}
\cmidrule(l){4-5}
& \textbf{Best baseline} & \textbf{L-FNO}
& \textbf{Best baseline} & \textbf{L-FNO} \\
\midrule
B1-Rare
& $0.157{\pm}0.012$ & $\mathbf{0.082{\pm}0.004}$
& $0.114{\pm}0.043$ & $\mathbf{0.750{\pm}0.021}$ \\

B2-Cascade
& $\mathbf{0.265{\pm}0.012}$ & $0.317{\pm}0.013$
& $0.790{\pm}0.022$ & $\mathbf{0.796{\pm}0.037}$ \\

B3-Burst
& $0.122{\pm}0.016$ & $\mathbf{0.104{\pm}0.016}$
& $0.246{\pm}0.173$ & $\mathbf{0.715{\pm}0.047}$ \\

B4-MultiScale
& $0.420{\pm}0.019$ & $\mathbf{0.297{\pm}0.039}$
& $0.368{\pm}0.165$ & $\mathbf{0.790{\pm}0.056}$ \\

B5-Nonlinear
& $0.400{\pm}0.015$ & $\mathbf{0.335{\pm}0.012}$
& $0.725{\pm}0.037$ & $\mathbf{0.866{\pm}0.013}$ \\

B6-Inhibitory
& $0.305{\pm}0.010$ & $\mathbf{0.188{\pm}0.009}$
& $0.257{\pm}0.046$ & $\mathbf{0.822{\pm}0.037}$ \\

B7-ZeroInflated
& $\mathbf{0.063{\pm}0.010}$ & $0.109{\pm}0.051$
& $0.262{\pm}0.157$ & $\mathbf{0.697{\pm}0.053}$ \\

B8-LongMemory
& $0.093{\pm}0.005$ & $\mathbf{0.088{\pm}0.011}$
& $0.078{\pm}0.037$ & $\mathbf{0.549{\pm}0.118}$ \\

\midrule
\textbf{Average}
& $0.231{\pm}0.019$ & $\mathbf{0.190{\pm}0.019}$
& $0.347{\pm}0.085$ & $\mathbf{0.748{\pm}0.048}$ \\
\bottomrule
\end{tabular*}

\vspace{0.25em}
\begin{minipage}{5.5in}
\scriptsize
\textit{Note.} Best baseline is selected between FNO-MSE and FNO-NLL separately for each scenario and metric. Values are mean$\pm$std over five random seeds.
\end{minipage}
\end{table}

\subsection{Real-World Benchmarks}

Table~\ref{tab:real_world_results} reports real-world benchmark results as mean $\pm$ standard deviation over random seeds. L-FNO achieves the best mean NLL, Brier score, PR-AUC, and AUC on all three datasets. Across datasets, L-FNO improves the average NLL from $0.186 \pm 0.012$ for FNO-NLL to $0.114 \pm 0.007$, and improves average PR-AUC from $0.271 \pm 0.040$ to $0.739 \pm 0.032$. The strongest discrimination gains appear in FDC and LSD, where L-FNO reaches PR-AUC values of $0.885 \pm 0.008$ and $0.772 \pm 0.012$, respectively. These results indicate that the Lorentzian spectral memory block provides a useful inductive bias for sparse stochastic event modeling beyond the likelihood objective alone.

\begin{table}[htbp]
\centering
\footnotesize
\setlength{\tabcolsep}{5pt}
\renewcommand{\arraystretch}{1.08}
\caption{Real-world benchmark summary over five random seeds. We compare L-FNO against the best baseline for each dataset and metric. Full mean$\pm$std results are reported in Appendix~\ref{app:full_results}.}
\label{tab:real_world_results}
\begin{tabular*}{5.5in}{@{\extracolsep{\fill}}lcccc@{}}
\toprule
\multirow{2}{*}{\textbf{Dataset}} &
\multicolumn{2}{c}{\textbf{NLL} $\downarrow$} &
\multicolumn{2}{c}{\textbf{PR-AUC} $\uparrow$} \\
\cmidrule(lr){2-3}
\cmidrule(l){4-5}
& \textbf{Best baseline} & \textbf{L-FNO}
& \textbf{Best baseline} & \textbf{L-FNO} \\
\midrule
D1-FDC
& $0.200{\pm}0.016$ & $\mathbf{0.134{\pm}0.011}$
& $0.475{\pm}0.022$ & $\mathbf{0.885{\pm}0.008}$ \\

D2-SECOM
& $0.265{\pm}0.016$ & $\mathbf{0.163{\pm}0.011}$
& $0.105{\pm}0.023$ & $\mathbf{0.559{\pm}0.077}$ \\

D3-LSD
& $0.092{\pm}0.005$ & $\mathbf{0.046{\pm}0.001}$
& $0.231{\pm}0.075$ & $\mathbf{0.772{\pm}0.012}$ \\

\midrule
\textbf{Average}
& $0.186{\pm}0.012$ & $\mathbf{0.114{\pm}0.007}$
& $0.271{\pm}0.040$ & $\mathbf{0.739{\pm}0.032}$ \\
\bottomrule
\end{tabular*}

\vspace{0.25em}
\begin{minipage}{5.5in}
\scriptsize
\textit{Note.} Best baseline is selected between FNO-MSE and FNO-NLL separately for each dataset and metric. Values are mean$\pm$std over five random seeds.
\end{minipage}
\end{table}

\begin{figure}[t]
  \centering
  \includegraphics[width=5.5in,height=1.2in]{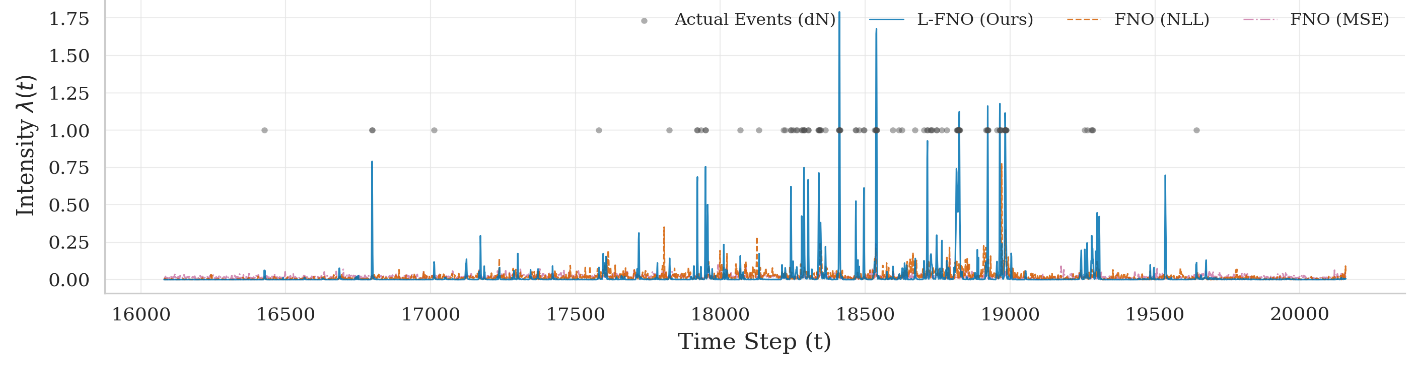}
  \caption{Model fitting and intensity estimation for LSD outbreak patterns in Thailand.}
  \label{fig:3}
  \vspace{0.25em}
  \begin{minipage}{5in}
    \scriptsize
    \textit{Note.} L-FNO captures nonlinear spikes in outbreak intensity with higher fidelity than baseline FNO models. While L-FNO demonstrates superior sensitivity to clustered outbreak events, it tends to overestimate during peak volatility, providing a conservative yet robust indicator for epidemic risk signaling.
  \end{minipage}
\end{figure}
\FloatBarrier

\section{Conclusion and Discussion}

The paper proposed L-FNO, bridging Fourier operator learning and temporal point processes via a structured spectral kernel and a Poisson-likelihood training scheme. The core insight is that standard regression objectives do not effectively define rare-event prediction problems, and resolving this misalignment yields substantial gains in event likelihood, calibration diagnostics, and detection performance across diverse benchmarks. The learned Lorentzian parameters provide a principled characterization of the history-dependent predictability structure of a stochastic event system, quantifying the strength and timescale of self-excitation without relying on domain-specific assumptions. This positions L-FNO not only as a predictive model but also as an analytical framework for assessing temporal structure under uncertainty.

L-FNO suggests a path from deterministic operator surrogates toward neural surrogates for stochastic counting processes and event-driven dynamics. By parameterizing the impulse response of self-exciting event processes through Lorentzian spectral kernels, L-FNO estimates conditional-intensity structure under sparsity. The model can reveal predictive temporal structure even under sparse observations, where standard regression-based operators often fail.

Limitations and future work include the following. First, the Lorentzian parameterization enforces \(\alpha>0\). This constraint precludes inhibitory dynamics, such as refractory behavior, where past events suppress future intensity. Transitioning to signed kernels would address the limitation. Second, the current spatial formulation treats regions as loosely coupled time series. A more principled approach would involve a fully spatiotemporal operator that jointly models space and time. Finally, extensions to longer temporal contexts and continuous-time formulations will widen the model's applicability to complex event-driven systems. L-FNO is intended as a decision-support tool for early risk detection. For high-stakes applications, L-FNO should be deployed only as decision support, with domain-expert review and uncertainty-aware interpretation to mitigate risks from false alarms, missed detections, and over-reliance on automated intensity estimates.
\bibliographystyle{unsrtnat}
\bibliography{references}

\clearpage
\appendix

\section{Reproducibility and Asset Documentation}
\label{app:reproducibility_assets}

To support reproducibility, we will release the implementation, synthetic data generators, baseline scripts, and public-data preprocessing code in an anonymized repository during review and make the repository public upon acceptance. Public datasets are acknowledged as originating from SECOM/UCI, ERA5/Copernicus, and WOAH-WAHIS. The proprietary FDC dataset cannot be redistributed due to access restrictions; however, we document its preprocessing pipeline, feature construction, train--test splitting procedure, and evaluation protocol to enable methodological scrutiny.

\section{Broader Impact Details}
\label{app:broader_impact_details}

L-FNO may aid early risk detection in public health, environmental monitoring, and industrial safety settings. However, false alarms may impose unnecessary operational costs, missed detections may delay intervention, and over-reliance on automated forecasts may reduce expert oversight. Accordingly, L-FNO should be deployed only as decision support, with domain-expert review, uncertainty-aware interpretation, and appropriate safeguards for high-stakes use cases.

\section{Proof of Proposition 1}
\label{app:proof}
\begin{quote}
\emph{\textbf{Proposition 1 (Approximation of causal kernels by stable exponential modes).} Let} 
\(g \in L^{2}\left([0,\infty)\right)\) 
\emph{be a causal finite-energy kernel. Then for any} 
\(\varepsilon > 0\), 
\emph{there exist an integer} \(K\), 
\emph{coefficients} 
\(\alpha_{1},\ldots,\alpha_{K}\in\mathbb{C}\), 
\emph{and decay rates} 
\(\beta_{1},\ldots,\beta_{K} > 0\) 
\emph{such that}
\[
\left\| 
g(t) - \sum_{k = 1}^{K}\alpha_{k}e^{-\beta_k t}
\right\|_{L^{2}\left([0,\infty)\right)} 
< \varepsilon.
\]
\emph{Equivalently, under the Paley--Wiener isometry, the corresponding stable rational transfer functions}
\[
\sum_{k = 1}^{K}\frac{\alpha_{k}}{\beta_{k} + s}
\]
\emph{are dense in} \(H^{2}(\mathbb{C}_{+})\), 
\emph{the Hardy space representation of causal} \(L^{2}\) 
\emph{kernels. Their boundary values on} \(s = i\omega\),
\[
\sum_{k = 1}^{K}\frac{\alpha_{k}}{\beta_{k} + i\omega},
\]
\emph{approximate the Fourier--Laplace boundary response of} \(g\) 
\emph{in} \(L^{2}(\mathbb{R}_{\omega})\). 
\emph{Each first-order pole has a Lorentzian magnitude-squared profile, while the complex response also encodes phase.}
\emph{When} \(g\) \emph{is real-valued, the coefficients can be chosen real-valued.}
\end{quote}

\emph{Proof.}

\paragraph{Step 1. Hardy space isomorphism.}

The one-sided Laplace transform 
\[
G(s)=\int_{0}^{\infty} g(t)e^{-st}dt
\]
maps \(L^{2}([0,\infty))\) isometrically, up to the usual Fourier normalization, onto \(H^2(\mathbb{C}_+)\), the space of analytic functions on the right half-plane bounded in the \(H^2\) sense. This is the Paley--Wiener theorem \citep{rudin1987realcomplex} for \(L^{2}([0,\infty))\): causal \(L^{2}\) functions correspond to boundary values of \(H^2(\mathbb{C}_+)\) on the imaginary axis \(s=i\omega\), understood in the \(L^2(\mathbb R_\omega)\) sense. Approximating \(g\) in \(L^{2}([0,\infty))\) is therefore equivalent to approximating \(G\) in \(H^2(\mathbb{C}_+)\).

\paragraph{Step 2. Density of rational functions in \(H^2(\mathbb{C}_+)\).}

Let 
\[
\mathcal R
=
\operatorname{span}_{\mathbb C}
\left\{
\frac{1}{\beta+s}:\beta>0
\right\}.
\]
Each summand \((\beta+s)^{-1}\) lies in \(H^2(\mathbb C_+)\) since its pole is in the left half-plane. It is a classical result \citep{garnett1981bounded} that the complex linear span of these Cauchy kernels is dense in \(H^2(\mathbb C_+)\). Equivalently, if \(F\in H^2(\mathbb C_+)\) is orthogonal to all \((\beta+s)^{-1}\), then the reproducing-kernel property implies
\[
F(\beta)=0
\qquad \forall \beta>0.
\]
Since \(F\) is analytic on \(\mathbb C_+\), the identity theorem gives \(F\equiv0\). Hence the orthogonal complement of \(\mathcal R\) is trivial, and \(\mathcal R\) is dense. Therefore, for any \(G\in H^2(\mathbb C_+)\) and \(\varepsilon>0\), there exists \(R_K\in\mathcal R\) such that
\[
\left\|G-R_K\right\|_{H^2}<\varepsilon.
\]

\paragraph{Step 3. Boundary values and time-domain equivalence.}

The boundary values of \(R_K(s)\) on \(s=i\omega\) are precisely
\[
R_K(i\omega)=\sum_{k=1}^{K}\frac{\alpha_k}{\beta_k+i\omega},
\]
the stable first-order rational frequency response used by the Lorentzian memory block. Since the Paley--Wiener isomorphism is an isometry, up to the chosen Fourier normalization, the \(H^2\) approximation error in Step 2 equals the \(L^{2}([0,\infty))\) error in the time domain. The inverse Fourier transform of
\[
\frac{\alpha_k}{\beta_k+i\omega}
\]
is
\[
\alpha_k e^{-\beta_k t}\mathbb{1}(t\geq 0).
\]
Therefore, the spectral parameterization used by L-FNO is equivalent to a finite exponential-mode representation of causal kernels in the time domain. The factor \(\mathbb{1}(t\geq 0)\) makes the causal support explicit. This completes the proof.

\paragraph{Remark A.1.} 
In L-FNO, the decay rates \(\{\beta_k\}\) are initialized through log-spaced timescales 
\[
\tau_k = 1/\beta_k \in \{1.25, 3, 10, 20\}
\]
and learned end-to-end. Proposition 1 should be interpreted as an asymptotic expressivity result for signed or complex amplitudes with positive decay rates: with sufficiently many trainable stable exponential modes, finite signed mixtures can approximate causal \(L^2\) excitation kernels. The implementation used in the main experiments constrains Lorentzian amplitudes to be nonnegative, matching the excitatory Hawkes setting. This positive-amplitude restriction does not cover arbitrary signed causal kernels and excludes inhibitory effects. Extending L-FNO to signed Lorentzian mixtures is a natural direction for modeling refractory or suppressive event dynamics.

\section{Implementation Details of the Lorentzian--Fourier Neural Operator}
\label{app:lfno_implementation}

This appendix provides the implementation-level details needed to reproduce the core L-FNO block. L-FNO takes an exogenous covariate tensor
\[
x \in \mathbb{R}^{B \times C_{\mathrm{in}} \times L}
\]
and a strictly lagged event-history sequence
\[
h \in \mathbb{R}^{B \times L},
\]
where \(B\) is the batch size, \(C_{\mathrm{in}}\) is the number of covariate channels, and \(L\) is the sequence length. The event-history input \(h\) is constructed only from past event indicators, so the current event \(dN(t)\) is not used when predicting \(\lambda(t)\).

The covariates are first embedded into a width-\(d\) latent representation,
\[
z^{(0)}(t)=\mathrm{Linear}(x(t)), 
\qquad z^{(0)}(t)\in\mathbb{R}^{d}.
\]
Each L-FNO block then applies two spectral paths. The free FNO path applies a learnable complex-valued operator to the Fourier transform of the latent covariates:
\[
\widehat{z}^{\,\mathrm{free}}_{b,o,m}
=
\sum_{i=1}^{d}
\widehat{z}_{b,i,m} W_{i,o,m},
\qquad m=1,\ldots,M_f .
\]
The Lorentzian event-history path transforms the lagged event-history sequence and applies a learnable Lorentzian spectral weighting:
\[
G_c(\omega_m)
=
\sum_{k=1}^{M_s}
\frac{\alpha_{c,k}}{\beta_{c,k}+i\omega_m},
\qquad
\widehat{h}^{\,\mathrm{lor}}_{b,c,m}
=
G_c(\omega_m)\widehat{h}_{b,m}.
\]
Here, \(\alpha_{c,k}=\operatorname{Softplus}(a_{c,k})\) and \(\beta_{c,k}=\operatorname{Softplus}(b_{c,k})\) ensure positive amplitude and decay parameters, and
\[
\omega_m = 2\pi\cdot \operatorname{rfftfreq}(L)_m .
\]
The two spectral paths are added and transformed back to the time domain:
\[
u^{(\ell)}
=
\mathcal{F}^{-1}
\left[
\widehat{z}^{\,\mathrm{free},(\ell)}
+
\widehat{h}^{\,\mathrm{lor},(\ell)}
\right].
\]
A \(1\times1\) skip connection and GELU nonlinearity produce the next block representation:
\[
z^{(\ell+1)}
=
\operatorname{GELU}
\left(
u^{(\ell)}
+
\operatorname{Conv1D}_{1\times1}(z^{(\ell)})
\right).
\]
The model stacks three such blocks and maps the final hidden representation to a nonnegative conditional intensity through a pointwise feed-forward head and Softplus activation:
\[
\widehat{\lambda}(t)
=
\operatorname{Softplus}
\left(
W_2\operatorname{GELU}(W_1z^{(3)}(t)+b_1)+b_2
\right).
\]

\subsection{Code Snippet}
\label{app:code_snippet}

The following snippet shows the core spectral update inside each L-FNO block. The full implementation, training scripts, synthetic generators, and preprocessing code will be released in an anonymized repository after review.

\begin{verbatim}
def lfno_block(x, h, omega):
    # x: latent covariates, shape (B, C, L)
    # h: strictly lagged event history, shape (B, L)

    xf = torch.fft.rfft(x)
    hf = torch.fft.rfft(h).expand(x.shape[0], x.shape[1], -1)

    # Free FNO path: R(omega) x_hat(omega)
    free_out = free_spectral_conv(xf)

    # Lorentzian event-history path
    alpha = F.softplus(log_alpha)
    beta = F.softplus(log_beta)

    om = omega.float().view(1, 1, -1).to(h.device)
    om_c = torch.complex(torch.zeros_like(om), om)

    G = alpha.unsqueeze(-1) / (beta.unsqueeze(-1) + om_c)
    hist_out = G.sum(1).unsqueeze(0) * hf

    # Merge spectral paths and return to time domain
    out = torch.fft.irfft(free_out + hist_out, n=x.shape[-1])

    return F.gelu(out + skip(x))
\end{verbatim}

\section{Full Multi-Seed Benchmark Results}
\label{app:full_results}

\subsection{Synthetic Benchmarks}
\label{app:synthetic_full_results}

Table~\ref{tab:app_synthetic_full_results} reports the full synthetic benchmark results over five random seeds as mean $\pm$ standard deviation. L-FNO achieves the highest PR-AUC in all eight synthetic scenarios, improving the macro-average PR-AUC from $0.347 \pm 0.085$ for FNO-NLL to $0.748 \pm 0.048$. It also obtains the best macro-average NLL, Brier score, and AUC. The gains are especially large in rare or structurally difficult regimes such as B1-Rare, B4-MultiScale, B6-Inhibitory, and B8-LongMemory, where the Lorentzian event-history path provides a strong inductive bias for sparse stochastic event dynamics.

The main exceptions are calibration-oriented metrics in selected regimes. FNO-NLL achieves the lowest NLL in B2-Cascade and B7-ZeroInflated, and the lowest Brier score in B2-Cascade, B7-ZeroInflated, and B8-LongMemory. This suggests that while the Lorentzian memory improves event ranking and rare-event discrimination, likelihood calibration can be more sensitive in regimes with frequent cascades, structurally silent periods, or long-memory dynamics.

\begin{table}[htbp]
\centering
\scriptsize
\setlength{\tabcolsep}{3pt}
\renewcommand{\arraystretch}{0.98}
\caption{Full synthetic benchmark results over five random seeds. Values are reported as mean $\pm$ standard deviation. Bold values indicate the best mean performance for each metric within each scenario.}
\label{tab:app_synthetic_full_results}
\begin{tabular*}{5.5in}{@{\extracolsep{\fill}}llcccc@{}}
\toprule
\textbf{Scenario} & \textbf{Model} &
\textbf{NLL} $\downarrow$ &
\textbf{Brier} $\downarrow$ &
\textbf{PR-AUC} $\uparrow$ &
\textbf{AUC} $\uparrow$ \\
\midrule

\multirow{3}{*}{B1-Rare}
& FNO-MSE & $0.172 \pm 0.008$ & $0.039 \pm 0.002$ & $0.077 \pm 0.033$ & $0.623 \pm 0.065$ \\
& FNO-NLL & $0.157 \pm 0.012$ & $0.036 \pm 0.001$ & $0.114 \pm 0.043$ & $0.700 \pm 0.082$ \\
& L-FNO & $\mathbf{0.082 \pm 0.004}$ & $\mathbf{0.018 \pm 0.001}$ & $\mathbf{0.750 \pm 0.021}$ & $\mathbf{0.975 \pm 0.003}$ \\

\addlinespace[0.2em]
\multirow{3}{*}{B2-Cascade}
& FNO-MSE & $0.278 \pm 0.012$ & $0.059 \pm 0.004$ & $0.781 \pm 0.029$ & $0.927 \pm 0.020$ \\
& FNO-NLL & $\mathbf{0.265 \pm 0.012}$ & $\mathbf{0.057 \pm 0.003}$ & $0.790 \pm 0.022$ & $\mathbf{0.935 \pm 0.013}$ \\
& L-FNO & $0.317 \pm 0.013$ & $0.086 \pm 0.009$ & $\mathbf{0.796 \pm 0.037}$ & $0.928 \pm 0.013$ \\

\addlinespace[0.2em]
\multirow{3}{*}{B3-Burst}
& FNO-MSE & $0.144 \pm 0.009$ & $0.032 \pm 0.003$ & $0.151 \pm 0.122$ & $0.728 \pm 0.047$ \\
& FNO-NLL & $0.122 \pm 0.016$ & $0.028 \pm 0.004$ & $0.246 \pm 0.173$ & $0.792 \pm 0.080$ \\
& L-FNO & $\mathbf{0.104 \pm 0.016}$ & $\mathbf{0.018 \pm 0.003}$ & $\mathbf{0.715 \pm 0.047}$ & $\mathbf{0.917 \pm 0.013}$ \\

\addlinespace[0.2em]
\multirow{3}{*}{B4-MultiScale}
& FNO-MSE & $0.420 \pm 0.019$ & $0.119 \pm 0.004$ & $0.254 \pm 0.053$ & $0.639 \pm 0.045$ \\
& FNO-NLL & $0.438 \pm 0.061$ & $0.114 \pm 0.014$ & $0.368 \pm 0.165$ & $0.710 \pm 0.101$ \\
& L-FNO & $\mathbf{0.297 \pm 0.039}$ & $\mathbf{0.056 \pm 0.010}$ & $\mathbf{0.790 \pm 0.056}$ & $\mathbf{0.926 \pm 0.028}$ \\

\addlinespace[0.2em]
\multirow{3}{*}{B5-Nonlinear}
& FNO-MSE & $0.400 \pm 0.015$ & $0.099 \pm 0.007$ & $0.692 \pm 0.042$ & $0.866 \pm 0.022$ \\
& FNO-NLL & $0.402 \pm 0.026$ & $0.095 \pm 0.010$ & $0.725 \pm 0.037$ & $0.880 \pm 0.023$ \\
& L-FNO & $\mathbf{0.335 \pm 0.012}$ & $\mathbf{0.062 \pm 0.004}$ & $\mathbf{0.866 \pm 0.013}$ & $\mathbf{0.948 \pm 0.008}$ \\

\addlinespace[0.2em]
\multirow{3}{*}{B6-Inhibitory}
& FNO-MSE & $0.311 \pm 0.010$ & $0.087 \pm 0.003$ & $0.243 \pm 0.044$ & $0.735 \pm 0.040$ \\
& FNO-NLL & $0.305 \pm 0.010$ & $0.086 \pm 0.003$ & $0.257 \pm 0.046$ & $0.767 \pm 0.036$ \\
& L-FNO & $\mathbf{0.188 \pm 0.009}$ & $\mathbf{0.042 \pm 0.004}$ & $\mathbf{0.822 \pm 0.037}$ & $\mathbf{0.965 \pm 0.008}$ \\

\addlinespace[0.2em]
\multirow{3}{*}{B7-ZeroInflated}
& FNO-MSE & $0.096 \pm 0.008$ & $0.017 \pm 0.002$ & $0.262 \pm 0.157$ & $0.870 \pm 0.062$ \\
& FNO-NLL & $\mathbf{0.063 \pm 0.010}$ & $\mathbf{0.014 \pm 0.002}$ & $0.201 \pm 0.159$ & $0.868 \pm 0.063$ \\
& L-FNO & $0.109 \pm 0.051$ & $0.020 \pm 0.013$ & $\mathbf{0.697 \pm 0.053}$ & $\mathbf{0.955 \pm 0.047}$ \\

\addlinespace[0.2em]
\multirow{3}{*}{B8-LongMemory}
& FNO-MSE & $0.114 \pm 0.007$ & $0.021 \pm 0.001$ & $0.021 \pm 0.003$ & $0.543 \pm 0.044$ \\
& FNO-NLL & $0.093 \pm 0.005$ & $\mathbf{0.019 \pm 0.000}$ & $0.078 \pm 0.037$ & $0.665 \pm 0.093$ \\
& L-FNO & $\mathbf{0.088 \pm 0.011}$ & $0.024 \pm 0.005$ & $\mathbf{0.549 \pm 0.118}$ & $\mathbf{0.953 \pm 0.014}$ \\

\midrule
\multirow{3}{*}{\textbf{Average}}
& \textbf{FNO-MSE} & $0.242 \pm 0.011$ & $0.059 \pm 0.003$ & $0.310 \pm 0.060$ & $0.741 \pm 0.043$ \\
& \textbf{FNO-NLL} & $0.231 \pm 0.019$ & $0.056 \pm 0.005$ & $0.347 \pm 0.085$ & $0.790 \pm 0.062$ \\
& \textbf{L-FNO} & $\mathbf{0.190 \pm 0.019}$ & $\mathbf{0.041 \pm 0.006}$ & $\mathbf{0.748 \pm 0.048}$ & $\mathbf{0.946 \pm 0.017}$ \\
\bottomrule
\end{tabular*}

\vspace{0.25em}
\begin{minipage}{5.5in}
\scriptsize
\textit{Note.} The average row reports the macro-average of scenario-level means and the macro-average of scenario-level standard deviations. NLL is the primary likelihood-based metric.
\end{minipage}
\end{table}

\subsection{Real-World Benchmarks}
\label{app:real_world_full_results}

Table~\ref{tab:app_real_world_full_results} reports the full real-world benchmark results over five random seeds as mean $\pm$ standard deviation. L-FNO achieves the best mean NLL, Brier score, PR-AUC, and AUC on all three datasets. Across datasets, L-FNO improves the average NLL from $0.186 \pm 0.012$ for FNO-NLL to $0.114 \pm 0.007$, and improves average PR-AUC from $0.271 \pm 0.040$ to $0.739 \pm 0.032$. The strongest discrimination gains appear in FDC and LSD, where L-FNO reaches PR-AUC values of $0.885 \pm 0.008$ and $0.772 \pm 0.012$, respectively. These results indicate that the Lorentzian spectral memory block provides a useful inductive bias for sparse stochastic event modeling beyond the likelihood objective alone.

\begin{table}[htbp]
\centering
\scriptsize
\setlength{\tabcolsep}{3pt}
\renewcommand{\arraystretch}{0.98}
\caption{Full real-world benchmark results over five random seeds. Values are reported as mean $\pm$ standard deviation. Bold values indicate the best mean performance for each metric within each dataset.}
\label{tab:app_real_world_full_results}
\begin{tabular*}{5.5in}{@{\extracolsep{\fill}}llcccc@{}}
\toprule
\textbf{Dataset} & \textbf{Model} &
\textbf{NLL} $\downarrow$ &
\textbf{Brier} $\downarrow$ &
\textbf{PR-AUC} $\uparrow$ &
\textbf{AUC} $\uparrow$ \\
\midrule

\multirow{3}{*}{D1-FDC}
& FNO-MSE & $0.240 \pm 0.010$ & $0.061 \pm 0.002$ & $0.408 \pm 0.029$ & $0.788 \pm 0.041$ \\
& FNO-NLL & $0.200 \pm 0.016$ & $0.051 \pm 0.004$ & $0.475 \pm 0.022$ & $0.844 \pm 0.023$ \\
& L-FNO & $\mathbf{0.134 \pm 0.011}$ & $\mathbf{0.025 \pm 0.003}$ & $\mathbf{0.885 \pm 0.008}$ & $\mathbf{0.987 \pm 0.001}$ \\

\addlinespace[0.2em]
\multirow{3}{*}{D2-SECOM}
& FNO-MSE & $0.300 \pm 0.011$ & $0.089 \pm 0.004$ & $0.078 \pm 0.008$ & $0.586 \pm 0.038$ \\
& FNO-NLL & $0.265 \pm 0.016$ & $0.078 \pm 0.009$ & $0.105 \pm 0.023$ & $0.662 \pm 0.062$ \\
& L-FNO & $\mathbf{0.163 \pm 0.011}$ & $\mathbf{0.043 \pm 0.005}$ & $\mathbf{0.559 \pm 0.077}$ & $\mathbf{0.934 \pm 0.016}$ \\

\addlinespace[0.2em]
\multirow{3}{*}{D3-LSD}
& FNO-MSE & $0.124 \pm 0.025$ & $0.024 \pm 0.001$ & $0.070 \pm 0.069$ & $0.614 \pm 0.093$ \\
& FNO-NLL & $0.092 \pm 0.005$ & $0.021 \pm 0.001$ & $0.231 \pm 0.075$ & $0.852 \pm 0.028$ \\
& L-FNO & $\mathbf{0.046 \pm 0.001}$ & $\mathbf{0.011 \pm 0.000}$ & $\mathbf{0.772 \pm 0.012}$ & $\mathbf{0.991 \pm 0.001}$ \\

\midrule
\multirow{3}{*}{\textbf{Average}}
& \textbf{FNO-MSE} & $0.221 \pm 0.015$ & $0.058 \pm 0.003$ & $0.186 \pm 0.035$ & $0.663 \pm 0.057$ \\
& \textbf{FNO-NLL} & $0.186 \pm 0.012$ & $0.050 \pm 0.005$ & $0.271 \pm 0.040$ & $0.786 \pm 0.038$ \\
& \textbf{L-FNO} & $\mathbf{0.114 \pm 0.007}$ & $\mathbf{0.026 \pm 0.003}$ & $\mathbf{0.739 \pm 0.032}$ & $\mathbf{0.971 \pm 0.006}$ \\
\bottomrule
\end{tabular*}

\vspace{0.25em}
\begin{minipage}{5.5in}
\scriptsize
\textit{Note.} The average row reports the macro-average of dataset-level means and the macro-average of dataset-level standard deviations. L-FNO outputs nonnegative Poisson intensity scores via Softplus. NLL is the primary likelihood-based metric.
\end{minipage}
\end{table}

\section{Additional Baselines: Details and Results}
\label{app:additional_baselines}
\subsection{Model Descriptions}

\textbf{FNO-WMSE.} Standard FNO trained with an aggressive event-upweighted MSE objective:
\[
\mathcal{L}_{\mathrm{WMSE}}
=
\frac{1}{L}
\sum_t
w_t
\left(
\widehat{\lambda}(t)-dN(t)
\right)^2,
\qquad
w_t
=
\left(
\frac{dN(t)}{\mathbb{E}[dN]}
\right)^{\gamma},
\qquad
\gamma=2.0.
\]
Because this weighting assigns zero loss weight to non-event timesteps, it is included as a diagnostic stress test for event-only reweighting rather than as a fully optimized calibrated baseline. Its role is to show that aggressively emphasizing rare positives can improve sensitivity at the cost of severe likelihood and probability-calibration degradation. The main calibrated operator baseline is therefore FNO-NLL, which shares the Poisson NLL objective with L-FNO but removes the structured Lorentzian memory block.

\textbf{Neural Hawkes (NH).} A continuous-time LSTM that models the conditional intensity from event history \(dN(t)\) alone, without any exogenous covariates. NH captures self-exciting dynamics but cannot incorporate high-dimensional covariate information \citep{mei2017neuralhawkes}.

\textbf{Neural Hawkes with Covariates (NH-X).} Extends NH \citep{isik2023flexiblehawkes,mei2017neuralhawkes} by concatenating \(X(t)\) to the LSTM input at each timestep. The sequential architecture limits capacity for covariate integration \citep{isik2023flexiblehawkes,xiao2019learning} across arbitrary temporal scales compared to L-FNO's parallel FNO operator stack \citep{li2021fno}.

\subsection{Shared Training Configuration}

Table~\ref{tab:hyperparameters} reports the training hyperparameters used for FNO models. We train all models with AdamW, using a weight decay of \(1\times10^{-4}\), an initial learning rate of \(3\times10^{-4}\), and cosine annealing over 300 epochs. Input windows have a length of 96 by default, except for D3-LSD, where \(L=64\). Sliding windows are generated with a stride of 8, except for LSD, where a stride of 4 is used. The FNO backbone has a width of 32. We retain 12 spectral modes in the covariate path and 4 modes in the Lorentzian path. Each Lorentzian block contains four Lorentzian terms, and three such blocks are stacked. The Lorentzian time-scale parameters are initialized as \(\tau\in\{1.25,3,10,20\}\), covering a range of short- to long-term temporal responses. Gradients are clipped at norm 1.0. The data are split chronologically into \(80\%\) training and \(20\%\) testing sets without shuffling. All experiments are repeated over five random seeds, and results are reported as mean \(\pm\) standard deviation.
\begin{table}[htbp]
\centering
\small
\setlength{\tabcolsep}{4pt}
\renewcommand{\arraystretch}{1.15}
\caption{Training parameters of FNO models.}
\label{tab:hyperparameters}
\begin{tabularx}{5.5in}{@{}p{1.55in}>{\raggedright\arraybackslash}X@{}}
\toprule
\textbf{Hyperparameter} & \textbf{Value} \\
\midrule
Optimizer & AdamW, with weight decay \(1\times 10^{-4}\) \\
Learning rate & \(3\times 10^{-4}\), with cosine annealing over 300 epochs \\
Sequence length & 96 timesteps; exception: \(L=64\) for D3-LSD \\
Sliding stride & 8; exception: 4 for D3-LSD \\
FNO width & 32 channels \\
Spectral modes & 12 for the covariate path; 4 for the Lorentzian path \\
Lorentzian terms & \(n=4\) per block, with 3 blocks \\
\(\tau\) initialization & \(\tau \in \{1.25, 3, 10, 20\}\), log-spaced \\
Gradient clipping & Maximum norm 1.0 \\
Train/test split & 80/20 chronological split, with no shuffle \\
Random seeds & Five independent seeds; results reported as mean \(\pm\) standard deviation \\
\bottomrule
\end{tabularx}
\end{table}

\subsection{Full Synthetic Results --- Appendix Models}

FNO-WMSE achieves NLL \(\approx 0.90\)--0.95 across all benchmarks as in Table~\ref{tab:synthetic-benchmarks}---near-random performance---confirming the miscalibration analysis of Section 3.4. NH and NH-X achieve competitive NLLs in low-rate scenarios but consistently underperform L-FNO on PR-AUC, reflecting the limitations of recurrent processing when rich covariate signals are available, and event detection rather than average log-likelihood is the primary criterion.

\begin{table}[htbp]
\centering
\caption{FNO-WMSE, NH, NH-X, and L-FNO* on synthetic benchmarks.}
\label{tab:synthetic-benchmarks}
\small
\setlength{\tabcolsep}{3.5pt}
\renewcommand{\arraystretch}{1.10}
\begin{tabular}{lcccccccc}
\toprule
& \multicolumn{4}{c}{NLL $\downarrow$}
& \multicolumn{4}{c}{PR-AUC $\uparrow$} \\
\cmidrule(lr){2-5}
\cmidrule(lr){6-9}
\textbf{Scenario}
& \textbf{FNO-WMSE}
& \textbf{NH}
& \textbf{NH-X}
& \textbf{L-FNO*}
& \textbf{FNO-WMSE}
& \textbf{NH}
& \textbf{NH-X}
& \textbf{L-FNO*} \\
\midrule
B1-Rare         & 0.9108 & 0.1772 & 0.1715 & \textbf{0.0795} & 0.1124 & 0.0406 & 0.0546 & \textbf{0.7654} \\
B2-Cascade      & 0.9532 & 0.3942 & 0.3992 & \textbf{0.3161} & 0.6951 & 0.3162 & 0.2585 & \textbf{0.8001} \\
B3-Burst        & 0.9378 & 0.1567 & 0.1546 & \textbf{0.1077} & 0.0663 & 0.0457 & 0.0521 & \textbf{0.7626} \\
B4-MultiScale   & 0.9530 & 0.3977 & 0.4116 & \textbf{0.2798} & 0.1703 & 0.0789 & 0.0935 & \textbf{0.6798} \\
B5-Nonlinear    & 0.9390 & 0.3885 & 0.3744 & \textbf{0.3332} & 0.5649 & 0.1744 & 0.2018 & \textbf{0.7803} \\
B6-Inhibitory   & 0.8943 & 0.2826 & 0.2910 & \textbf{0.1798} & 0.3902 & 0.0822 & 0.0965 & \textbf{0.6142} \\
B7-ZeroInflated & 0.9512 & 0.1764 & 0.1872 & \textbf{0.1181} & 0.3165 & 0.0578 & 0.0621 & \textbf{0.7642} \\
B8-LongMemory   & 0.9185 & 0.2241 & 0.2198 & \textbf{0.0807} & 0.1389 & 0.0516 & 0.0613 & \textbf{0.4812} \\
\bottomrule
\end{tabular}

\vspace{0.4em}
\begin{minipage}{5.5in}
\footnotesize
\textit{Note.}
L-FNO* denotes the proposed model. The table reports additional baseline results for FNO-WMSE, Neural Hawkes (NH), and Neural Hawkes with covariates (NH-X). The L-FNO* values are aligned with the main synthetic benchmark results in Table~\ref{tab:synthetic_results}.
\end{minipage}
\end{table}

\section{Synthetic Benchmark Design}

\subsection{Scenario Parameters and Stability Conditions}

Each scenario is generated via a discrete-time closed-loop Hawkes simulation in which
\[
dN_t \sim \mathrm{Bernoulli}(p_t), 
\qquad 
p_t=\operatorname{clip}(\lambda_t,\varepsilon,0.99).
\]
The history state is updated only after sampling the current event:
\[
h_{t+1}=e^{-\beta}h_t+dN_t.
\]
The intensity used to generate \(dN_t\) depends on \(h_t\), which contains events only up to time \(t-1\). Thus, the simulator is causal and does not use \(dN_t\) when computing the intensity for the same timestep. For scenarios with \(\alpha>0\), the discrete branching ratio \(|\eta|_{\mathrm{disc}}=\alpha/(1-e^{-\beta})\) satisfies \(|\eta|_{\mathrm{disc}}<1\), ensuring stationarity. B6-Inhibitory uses \(\alpha<0\) and B7-ZeroInflated uses a Markov switching mechanism.

\begin{table}[htbp]
\centering
\small
\setlength{\tabcolsep}{2pt}
\renewcommand{\arraystretch}{1.12}
\caption{Generative parameters for all eight synthetic scenarios.}
\label{tab:generative_parameters}
\begin{tabularx}{5.5in}{@{}l
>{\centering\arraybackslash}p{0.35in}
>{\centering\arraybackslash}p{0.58in}
>{\centering\arraybackslash}p{0.58in}
>{\centering\arraybackslash}p{0.55in}
>{\centering\arraybackslash}p{0.55in}
>{\centering\arraybackslash}p{0.40in}
>{\raggedright\arraybackslash}X@{}}
\toprule
\textbf{Scenario} &
\textbf{Rate} &
\(\boldsymbol{\alpha}\) &
\(\boldsymbol{\beta}\) &
\(\boldsymbol{\tau}\) &
\(\boldsymbol{|\eta|_{\mathrm{disc}}}\) &
\textbf{Fano} &
\textbf{Description} \\
\midrule
B1-Rare
& 0.039 & 0.30 & 1.00 & 1.0 & 0.47 & 2.65
& Sparse events with weak self-excitation \\
B2-Cascade
& 0.141 & 0.32 & 0.50 & 2.0 & 0.81 & 11.95
& Near-critical cascade dynamics \\
B3-Burst
& 0.041 & 0.35 & 0.70 & 1.4 & 0.69 & 9.15
& Intermittent ON/OFF regime with Hawkes excitation \\
B4-MultiScale
& 0.072 & \((0.30,0.02)\) & \((1.00,0.05)\) & \((1,20)\) & 0.88 & 8.30
& Dual kernel with fast and slow excitation components \\
B5-Nonlinear
& 0.162 & 0.30 & 0.60 & 1.7 & 0.66 & 4.95
& Nonlinear covariate interaction \(x_1 \times x_2\) \\
B6-Inhibitory
& 0.095 & -0.25 & 0.80 & 1.3 & --- & 0.82
& Inhibitory refractory dynamics \\
B7-ZeroInflated
& 0.046 & 0.35 & 0.60 & 1.7 & 0.78 & 13.97
& Markov switching with structural silence \\
B8-LongMemory
& 0.027 & \((0.20,0.006)\) & \((1.00,0.02)\) & \((1,50)\) & 0.62 & 2.17
& Dual kernel with fast and long-memory excitation components \\
\bottomrule
\end{tabularx}

\vspace{0.4em}
\begin{minipage}{5.5in}
\footnotesize
\textit{Note.}
\(\alpha\) denotes the excitation amplitude, which scales the Hawkes history contribution to the intensity. 
\(\beta\) is the decay rate of the exponential kernel \(e^{-\beta t}\), and \(\tau=1/\beta\) is the corresponding memory timescale. 
For a single excitatory exponential kernel, the discrete-time branching ratio is 
\(|\eta|_{\mathrm{disc}}=|\alpha|/(1-e^{-\beta})\). 
For additive multi-kernel scenarios such as B4-MultiScale and B8-LongMemory, we report the total branching ratio 
\(|\eta|_{\mathrm{disc}}=\sum_j |\alpha_j|/(1-e^{-\beta_j})\). 
Stationarity requires \(|\eta|_{\mathrm{disc}}<1\) for excitatory Hawkes dynamics. 
B6-Inhibitory is excluded from the branching-ratio interpretation because its negative feedback does not admit the usual offspring-process interpretation.
\end{minipage}
\end{table}

\section{Real-World Dataset Details}

\subsection{D1-FDC: Semiconductor Fault Detection}

Continuous process signals \(Y(t)\) from a semiconductor manufacturing process (etching process) over multiple production runs are used. Up to 85 process variables are extracted per timestep; missing timestamps are interpolated linearly. Event labels are derived via \(dN(t)=\mathbf{1}[Y(t)>\text{90th quantile of }Y]\), flagging the top 10\% of process excursions. Chronological 80/20 split is applied.

\subsection{D2-SECOM: Semiconductor Wafer Defect}

The UCI SECOM dataset contains 590 sensor measurements for 1,567 production runs. Sensors with more than 50\% missing values are discarded; remaining NaN values are median-imputed. Binary defect labels are provided directly.

\subsection{D3-LSD: Thailand LSD Outbreak 2021}

Province-level Lumpy Skin Disease outbreak records for all 67 Thai provinces over 365 days in 2021 are used. ERA5 climate reanalysis covariates, including temperature, humidity, wind, and precipitation, are aligned per province-day. A wind-directed spatial Hawkes kernel weights neighboring provinces by wind bearing and distance:
\[
K_{\mathrm{wind}}(p,p',t)
=
\exp\left(-\frac{d_{pp'}}{200}\right)
\cdot
\exp\left(\gamma \cos(\theta_{\mathrm{wind}}-\theta_{p\rightarrow p'})\right).
\]
The spatial history is computed causally as
\[
h_{\mathrm{wind}}(p,t)
=
\sum_{p'}
\sum_{s<t}
K_{\mathrm{wind}}(p,p',s)
dN(p',s)
e^{-\beta(t-s)}.
\]
This construction uses only past outbreak events and therefore does not leak the current label \(dN(p,t)\).

\section{Computational Resources}

All experiments were conducted on a single workstation with an Apple M-series processor (unified memory architecture). No cloud computing or multi-GPU training was used. Approximate runtimes per model per scenario at 300 epochs are summarized in Table~\ref{tab:computational_times}.

\begin{table}[htbp]
\centering
\small
\setlength{\tabcolsep}{4pt}
\renewcommand{\arraystretch}{1.12}
\caption{Computational times for benchmarks.}
\label{tab:computational_times}
\begin{tabularx}{5.5in}{@{}>{\raggedright\arraybackslash}p{2in}
>{\centering\arraybackslash}p{0.85in}
>{\raggedright\arraybackslash}X@{}}
\toprule
\textbf{Setting} &
\textbf{\shortstack{Time per model}} &
\textbf{Full suite} \\
\midrule
Synthetic B1--B8 \((T=5{,}000, L=96)\)
& \(\sim\)1 min
& \(\sim\)5 min per scenario; \(\sim\)40 min total \\
Real-world D1/D2 \((T\sim4{,}000)\)
& \(\sim\)3 min
& \(\sim\)20 min per dataset; \(\sim\)40 min total \\
Real-world D3-LSD \((T=24{,}455)\)
& \(\sim\)5 min
& \(\sim\)30 min \\
Full experiment, 8 synthetic + 3 real
& ---
& \(\sim\)2 hours \\
\bottomrule
\end{tabularx}
\end{table}

The code is implemented in pure PyTorch without custom CUDA kernels. A standard GPU would reduce the full suite's runtime substantially.

\section{Full Benchmark Visualizations}
\label{app:visualizations}
\subsection{Synthetic Benchmark Visualizations}

\begin{figure}[!htbp]
  \centering
  \includegraphics[width=\linewidth]{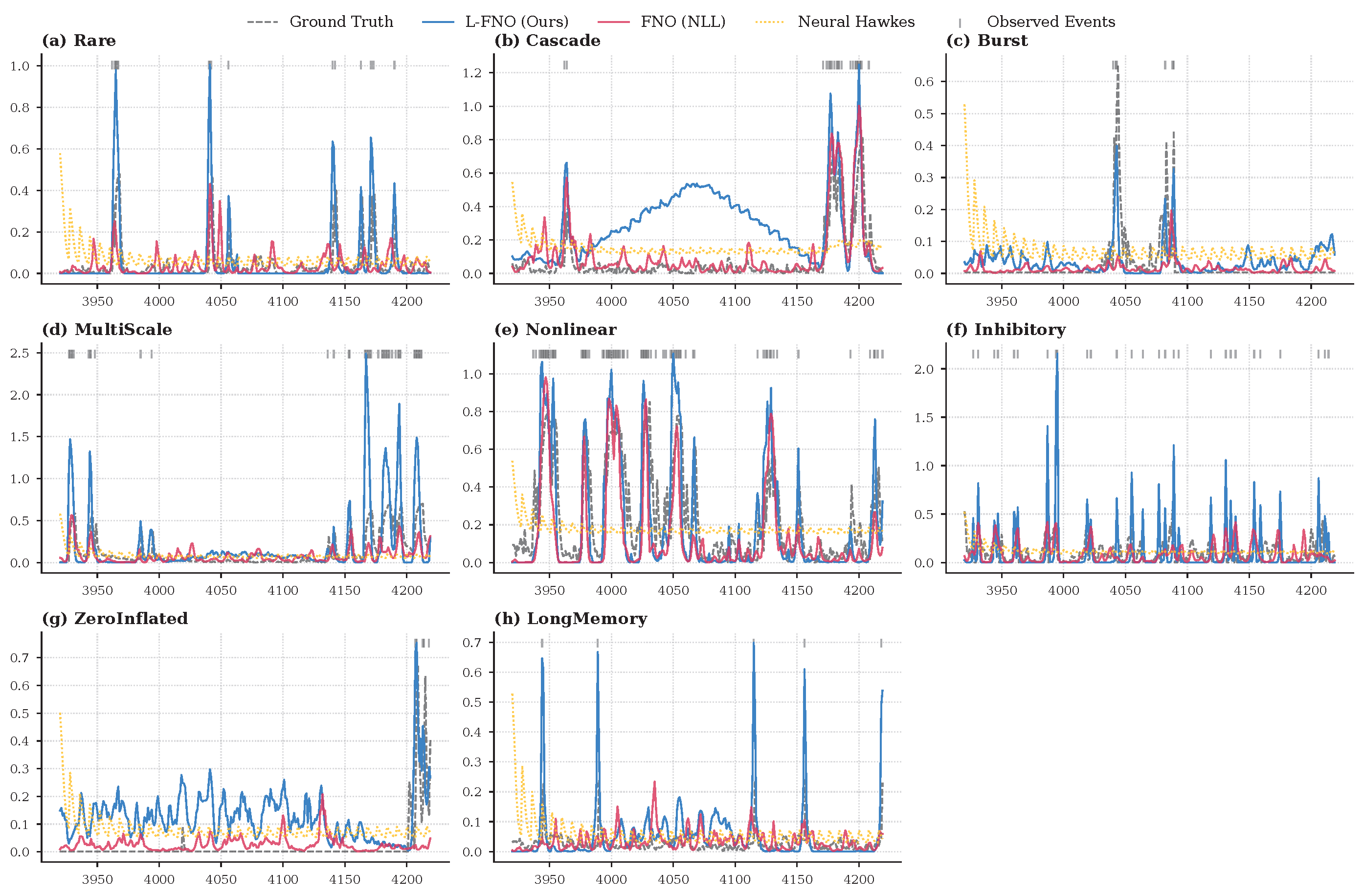}
  \caption{Comparative analysis of intensity estimation across synthetic benchmarks.}
  \label{fig:4}
\end{figure}

Figure~\ref{fig:4} illustrates the predicted conditional intensities \(\lambda(t)\) for eight distinct point process dynamics. Proposed L-FNO consistently captures the complex temporal dependencies and sharp transitions in benchmarks such as Burst and Cascade, outperforming the standard FNO-NLL and Neural Hawkes. Vertical ticks at the top of each subplot indicate observed event occurrences.

\clearpage
\subsection{Real-world Data Visualizations}

\begin{figure}[!htbp]
  \centering
  \includegraphics[width=\linewidth]{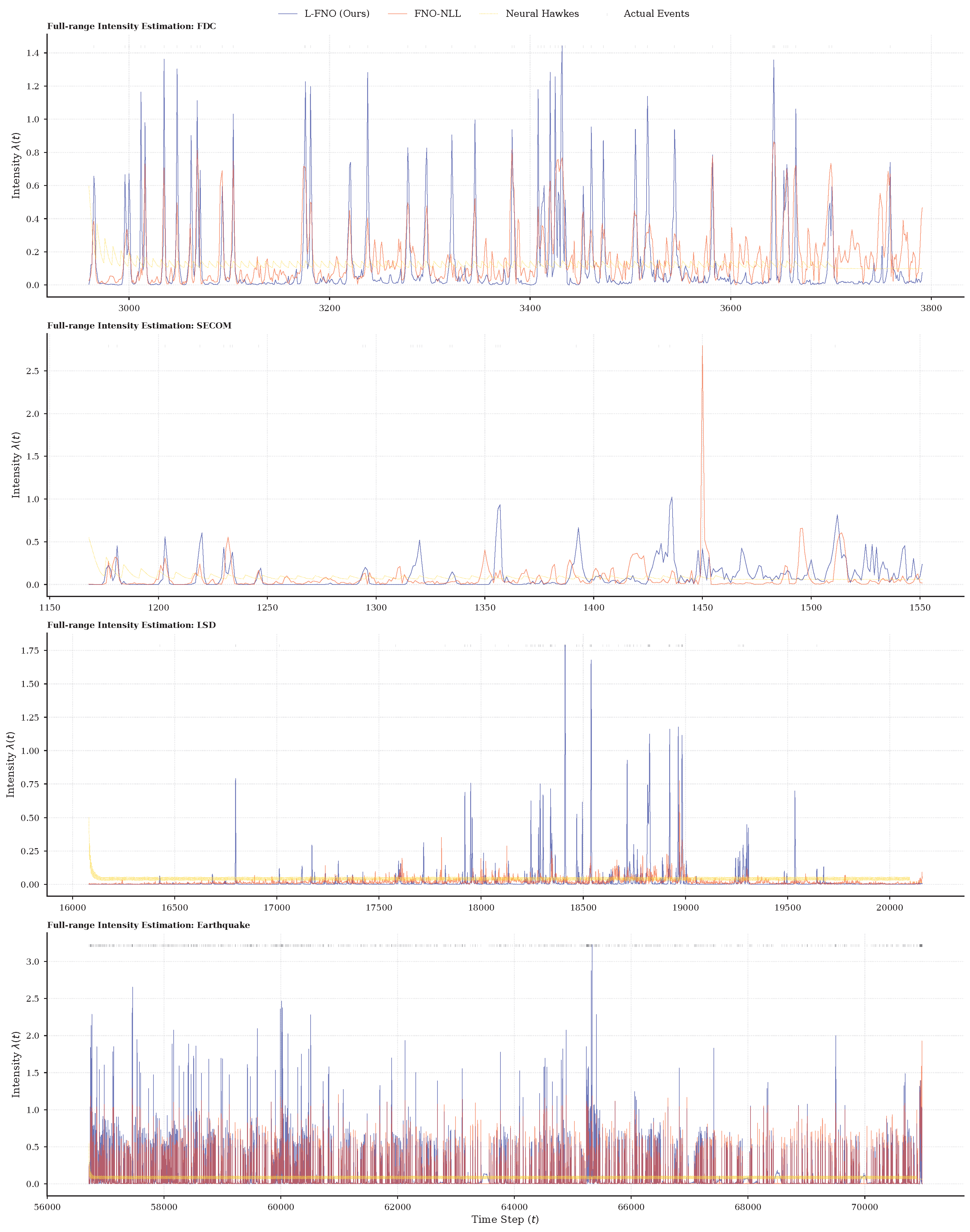}
  \caption{Comprehensive intensity estimation for real-world benchmarks.}
  \label{fig:5}
\end{figure}

Figure~\ref{fig:5} compares L-FNO, FNO-NLL, and Neural Hawkes across three real-world scenarios. L-FNO shows higher sensitivity to clustered event occurrences while maintaining a calibrated baseline during quiescent periods. In the LSD case, Neural Hawkes exhibits poor calibration and fails to adapt to climate-driven intensity changes, highlighting the limitation of purely history-dependent models when exogenous drivers dominate event dynamics.

\end{document}